\documentclass[11pt,a4paper]{article}
\usepackage[draft]{hyperref}
\usepackage{acl2018}
\usepackage{times}
\usepackage{latexsym}
\usepackage{xspace}
\usepackage{arydshln}
\usepackage{graphicx}
\usepackage{multirow}
\usepackage{amsmath}
\usepackage{relsize}
\usepackage{url}
\usepackage{caption}
\usepackage{subcaption}
\usepackage{adjustbox,booktabs,enumitem}
\usepackage{amsfonts,amssymb}
\usepackage{algorithm,algpseudocode}
\usepackage{scalerel}

\aclfinalcopy 
\def\aclpaperid{1089}

\newcommand{\method}[1]{\texttt{#1}\xspace}
\newcommand{\lm}{\method{LM}}
\newcommand{\lma}{\method{LM$^*$}}
\newcommand{\lmb}{\method{LM$^{**}$}}
\newcommand{\lmbconv}{\method{LM$^{**}$-C}}
\newcommand{\lmbpmrm}{\method{LM$^{**}$+PM+RM}}
\newcommand{\lmbrm}{\method{LM$^{**}$+RM}}
\newcommand{\human}{\method{Human}}
\newcommand{\stressb}{\method{Stress-BL}}
\newcommand{\rhymeb}{\method{Rhyme-BL}}
\newcommand{\rhymeem}{\method{Rhyme-EM}}

\newcommand{\sd}[1]{\small$\pm${#1}\xspace}

\newcommand{\dataset}[1]{\textsc{#1}\xspace}
\newcommand{\background}{\dataset{background}}
\newcommand{\sonnet}{\dataset{sonnet}}

\newcommand{\smallurl}[1]{{\smaller{\url{#1}}}}

\newcommand{\secref}[2][]{Section#1~\ref{sec:#2}}

\newcommand{\tabref}[2][]{Table#1~\ref{tab:#2}}
\newcommand{\figref}[2][]{Figure#1~\ref{fig:#2}}

\newcommand{\eqnref}[2][]{Equation#1~(\ref{eqn:#2})}

\newcommand{\ex}[1]{\textit{#1}\xspace}
\newcommand{\pair}[2]{(\ex{#1}, \ex{#2})}

\newcommand\email{\begingroup \urlstyle{tt}\smaller\Url}

\newcommand{\cev}[1]{\reflectbox{\ensuremath{\vec{\reflectbox{\ensuremath{#1}}}}}}

\title{Deep-speare: A joint neural model of poetic language, meter and 
rhyme}

\author{Jey Han Lau$^{1,2}$ \qquad Trevor
Cohn$^{2}$ \qquad Timothy Baldwin$^{2}$ \\ \textbf{Julian Brooke}$^{3}$ 
\qquad \textbf{Adam Hammond}$^{4}$ \\[1ex]
    $^1$ IBM Research Australia \\ $^2$ School of Computing and 
    Information Systems, The University of
Melbourne \\$^3$ Thomson Reuters \\ $^4$ Department of English, University of 
    Toronto \\[1ex]
    \email{jeyhan.lau@gmail.com},
\email{t.cohn@unimelb.edu.au}, \email{tb@ldwin.net},\\ 
\email{julian.brooke@gmail.com}, \email{adam.hammond@utoronto.ca}}

\date{}

\begin{document}

\maketitle

\begin{abstract}

In this paper, we propose a joint architecture 
that captures language, rhyme and meter for sonnet modelling.  We assess 
the quality of generated poems using crowd and expert judgements.  The 
stress and rhyme models perform very well, as generated poems are 
largely indistinguishable from human-written poems.  Expert evaluation, 
however, reveals that a vanilla language model captures meter implicitly, 
and that machine-generated poems still underperform in terms of 
readability and emotion. Our research shows the importance expert 
evaluation for poetry generation, and that future research should look 
beyond rhyme/meter and focus on poetic language.

\end{abstract}

\section{Introduction}

With the recent surge of interest in deep learning, one question that is
being asked across a number of fronts is: can deep learning techniques
be harnessed for creative purposes?  Creative applications where such
research exists include the composition of music
\cite{Humphrey+:2013,Sturm+:2016,Choi+:2016}, the design of
sculptures \cite{Lehman+:2016}, and automatic choreography
\cite{Crnkovic-Friis+:2016}. In this paper, we focus on a creative
textual task: automatic poetry composition.

A distinguishing feature of poetry is its \textit{aesthetic forms}, e.g.\
rhyme and rhythm/meter.\footnote{Noting that there are many notable 
divergences
  from this in the work of particular poets (e.g.\ Walt Whitman) and 
  poetry types (such as
  free verse or haiku).} In this work, we treat the task of poem 
generation as a constrained language modelling task, such that lines of 
a given poem rhyme, and each line follows a canonical meter and has a 
fixed number of stresses. Specifically, we focus on sonnets and generate 
quatrains in iambic pentameter (e.g.\ see \figref{sonnet18}), based on 
an unsupervised model of language, rhyme and meter
trained on a novel corpus of sonnets.

\begin{figure}[t]
  \centering
  \begin{minipage}[t]{0.95\linewidth}
    \ex{Shall I compare thee to a summer's day?} \\
    \ex{Thou art more lovely and more temperate:}\\
    \ex{Rough winds do shake the darling buds of May,}\\
    \ex{And summer's lease hath all too short a date:}
  \end{minipage}
\caption{1st quatrain of Shakespeare's \textit{Sonnet 18}.}
\label{fig:sonnet18}
\end{figure}

Our findings are as follows:
\begin{itemize}[nosep,leftmargin=1em,labelwidth=*,align=left]
\item our proposed stress and rhyme models work very well, generating
  sonnet quatrains with stress and rhyme patterns that are
  indistinguishable from human-written poems and rated highly by an
  expert;
\item a vanilla language model trained over our sonnet corpus,
  surprisingly, captures meter implicitly at human-level performance;
\item while crowd workers rate the poems generated by our best model as
  nearly indistinguishable from published poems by humans, an expert
  annotator found the machine-generated poems to lack readability and
  emotion, and our best model to be only comparable to a
  vanilla language model on these dimensions;
\item most work on poetry generation focuses on meter 
  \cite{Greene+:2010,Ghazvininejad+:2016,Hopkins+:2017}; our results 
  suggest that future research should look beyond meter and focus on 
  improving readability.
\end{itemize}
In this, we develop a new annotation framework for the evaluation of
machine-generated poems, and release both a novel data of sonnets
and the full source code associated with this
research.\footnote{\url{https://github.com/jhlau/deepspeare}}

\section{Related Work}

Early poetry generation systems were generally rule-based, and based on
rhyming/TTS dictionaries and syllable counting
\cite{Gervas:2000,Wu+:2009,Netzer+:2009,Colton+:2012,Toivanen+:2013}.
The earliest attempt at using statistical modelling for poetry
generation was \newcite{Greene+:2010}, based on a language model paired
with a stress model. 

Neural networks have dominated recent research. \newcite{Zhang+:2014}
use a combination of convolutional and recurrent networks for modelling
Chinese poetry, which \newcite{Wang+:2016} later simplified by
incorporating an attention mechanism and training at the character
level. For English poetry, \newcite{Ghazvininejad+:2016}
introduced a finite-state acceptor to explicitly model rhythm in
conjunction with a recurrent neural language model for generation.
\newcite{Hopkins+:2017} improve rhythm modelling with a cascade of
weighted state transducers, and demonstrate the use of character-level
language model for English poetry.  
A critical difference over our work is that we jointly model both poetry  
content and forms, and unlike previous work which use dictionaries 
\cite{Ghazvininejad+:2016} or heuristics \cite{Greene+:2010} for rhyme,
we learn it automatically.


\section{Sonnet Structure and Dataset}
\label{sec:data}

The sonnet is a poem type popularised by Shakespeare, made up of 14
lines structured as 3 quatrains (4 lines) and a couplet (2
lines);\footnote{There are other forms of sonnets, but the Shakespearean
  sonnet is the dominant one. Hereinafter ``sonnet'' is used to
  specifically mean Shakespearean sonnets.} an example quatrain is
presented in \figref{sonnet18}. It follows a number of \textit{aesthetic
forms}, of which two are particularly salient: stress and rhyme.

A sonnet line obeys an alternating stress pattern, called the iambic 
pentameter, e.g.:\\[-4ex]
\begin{center}
\begin{tabular}{c@{\;}c@{\;}c@{}c@{\;}c@{\;\;}c@{\;\;}c@{\;}c@{}c@{\;}c}
$S^-$ & $S^+$ & $S^-$ & $S^+$ & $S^-$ & $S^+$ & $S^-$ & $S^+$ & $S^-$ & 
 $S^+$\\
\ex{Shall} & \ex{I} & \ex{com} & \ex{pare} & \ex{thee} & \ex{to} & 
\ex{a} & \ex{sum} & \ex{mer's} & \ex{day?} 
\end{tabular}
\end{center}
where $S^-$ and $S^+$ denote unstressed and stressed syllables,
respectively.

A sonnet also rhymes, with a typical rhyming scheme being \textit{ABAB CDCD 
EFEF GG}.  There are a number of variants, however, mostly seen in 
the quatrains; e.g.\  \textit{AABB} or \textit{ABBA} are also common.

We build our sonnet dataset from the latest image of Project
Gutenberg.\footnote{\url{https://www.gutenberg.org/}.}  We first create
a (generic) poetry document collection using the GutenTag tool
\cite{GutenTag}, based on its inbuilt poetry classifier and rule-based
structural tagging of individual
poems. 

Given the poems, we use word and character statistics derived from
Shakespeare's 154 sonnets to filter out all non-sonnet poems (to form
the ``\background'' dataset), leaving the sonnet corpus
(``\sonnet'').\footnote{The following constraints were used to select
  sonnets: $8.0 \leqslant $ mean words per line $\leqslant 11.5$;
  $40 \leqslant $ mean characters per line $\leqslant 51.0$; min/max
  number of words per line of 6/15; min/max number of characters per
  line of 32/60; and min letter ratio per line $\geqslant 0.59$.} Based
on a small-scale manual analysis of \sonnet, we find that the approach
is sufficient for extracting sonnets with high precision. \background
serves as a large corpus (34M words) for pre-training word embeddings,
and \sonnet is further partitioned into training, development and
testing sets. Statistics of \sonnet are given in 
\tabref{dataset}.\footnote{The sonnets in our collection are largely in 
Modern English, with possibly a small number of poetry in Early Modern 
English. The potentially mixed-language dialect data might add noise to 
our system, and given more data it would be worthwhile to include time 
period as a factor in the model.}

\begin{table}[t]
\begin{center}
\begin{tabular}{ccc}
\toprule
\textbf{Partition} & \textbf{\#Sonnets} & \textbf{\#Words} \\
\midrule
Train & 2685 & 367K \\
Dev & 335 & 46K \\
Test & 335 & 46K \\
\bottomrule
\end{tabular}
\end{center}
\caption{\sonnet dataset statistics.}
\label{tab:dataset}
\end{table}

\section{Architecture}

\begin{figure*}[t]
        \centering
        \begin{subfigure}[b]{0.4\textwidth}
            \includegraphics[width=\textwidth]{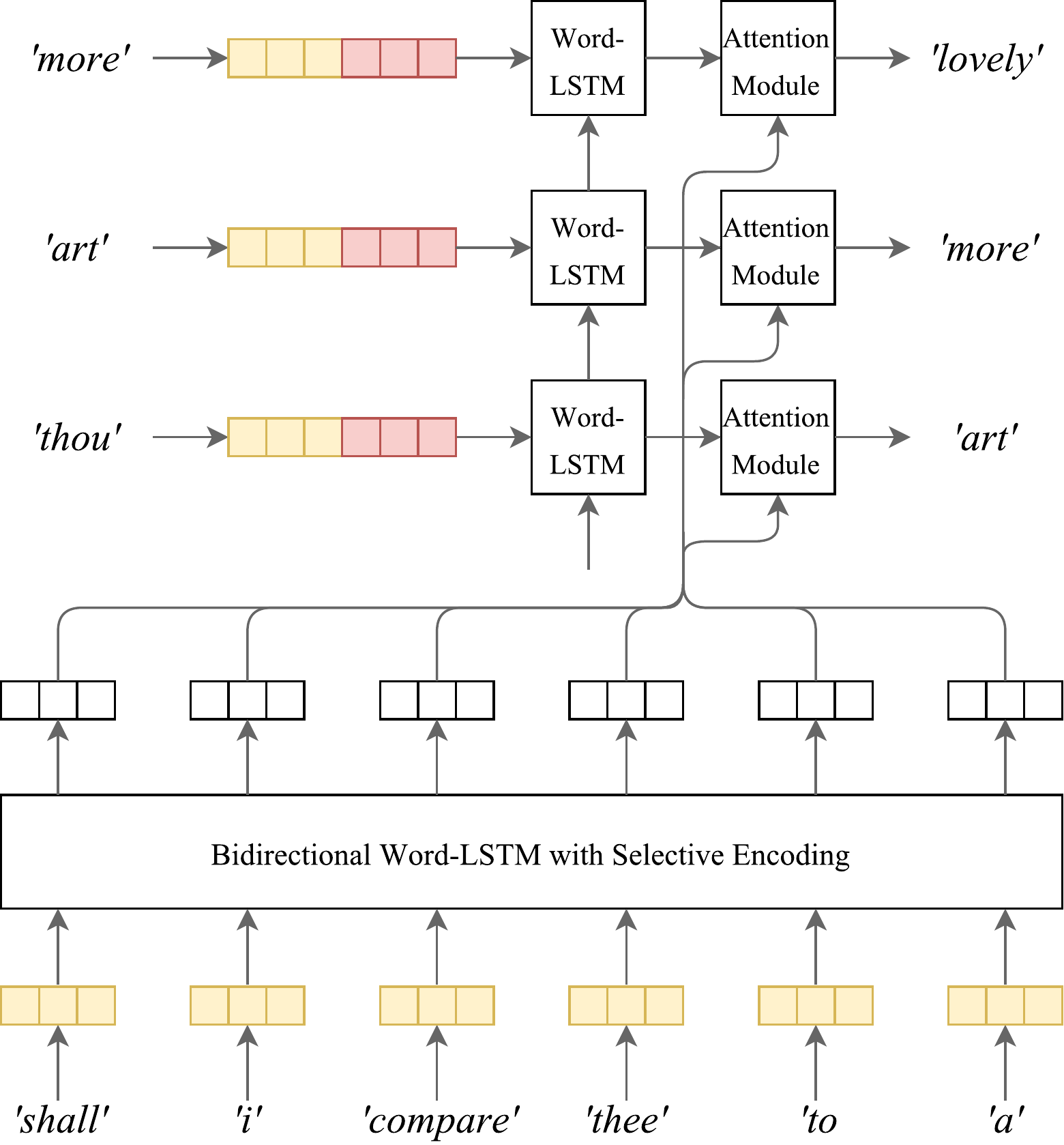}
            \caption{Language model}
        \end{subfigure}
        ~
        \begin{subfigure}[b]{0.45\textwidth}
            \includegraphics[width=\textwidth]{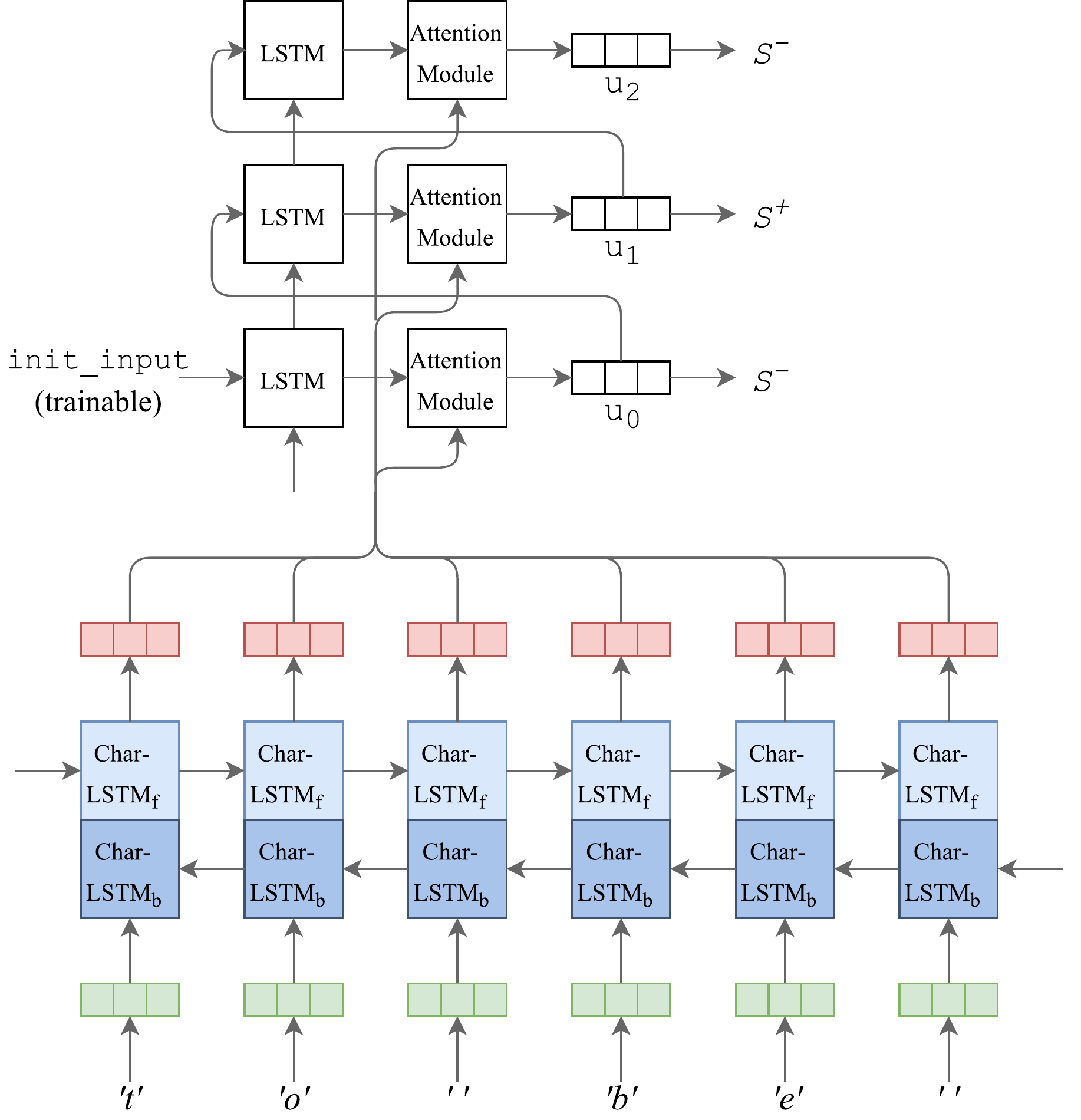}
            \caption{Pentameter model}
        \end{subfigure}
        ~
        \begin{subfigure}[b]{0.45\textwidth}
            \includegraphics[width=\textwidth]{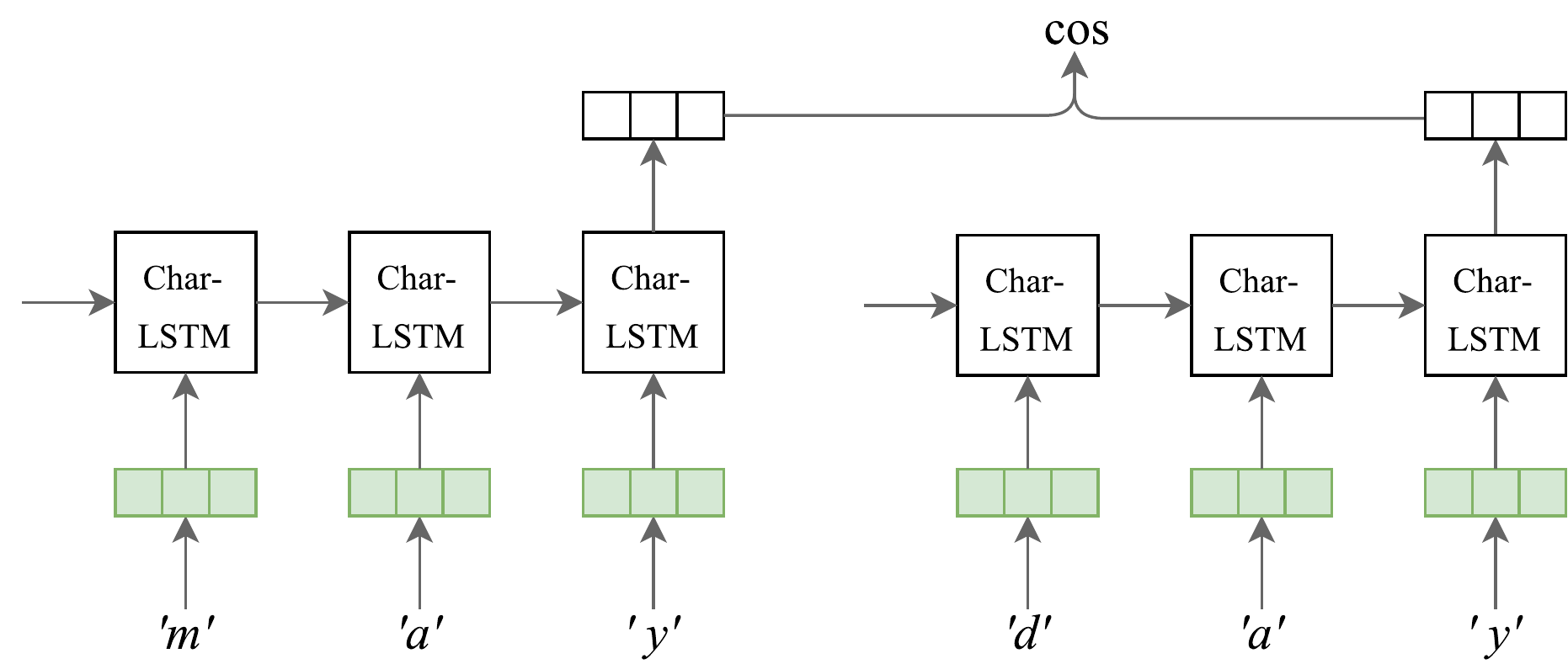}
            \caption{Rhyme model}
        \end{subfigure}
\caption{Architecture of the language, pentameter and rhyme models.  
Colours denote shared weights.}
\label{fig:architecture}
\end{figure*}

We propose modelling both content and forms jointly with a neural
architecture, composed of 3 components: (1) a language model; (2) a
pentameter model for capturing iambic pentameter; and (3) a rhyme model
for learning rhyming words.

Given a sonnet line, the language model uses standard categorical
cross-entropy to predict the next word, and the pentameter model
is similarly trained to learn the alternating iambic stress
patterns.\footnote{There are a number of variations in addition to the
  standard pattern \cite{Greene+:2010}, but our model uses only the
  standard pattern as it is the dominant one.} The rhyme model, on the
other hand, uses a margin-based loss to separate rhyming word pairs from
non-rhyming word pairs in a quatrain. For generation we use the language
model to generate one word at a time, while applying the
pentameter model to sample meter-conforming sentences and the rhyme
model to enforce rhyme. The architecture of the joint model is
illustrated in \figref{architecture}. We train all the components 
together by treating each component as a sub-task in a multi-task 
learning setting.\footnote{We stress that although the components appear 
to be disjointed, the shared parameters allow the components to mutually 
influence each other during joint training. To exemplify this, we found 
that the pentameter model performs very poorly when we train each 
component separately.}

\subsection{Language Model}
\label{sec:lm}

The language model is a variant of an LSTM encoder--decoder model with 
attention \cite{Bahdanau+:2015}, where the encoder encodes the preceding 
context (i.e.\ all sonnet lines before the current line) and the 
decoder decodes one word at a time for the current line, while attending 
to the preceding context.

In the encoder, we embed context words $z_i$ using embedding matrix
$\mathbf{W}_{wrd}$ to yield $\mathbf{w}_i$, and feed them to a
biLSTM\footnote{We use a single layer for all LSTMs.} to
produce a sequence of encoder hidden states
$\mathbf{h}_i = [\vec{\mathbf{h}}_i; \cev{\mathbf{h}}_i]$.  Next we
apply a selective mechanism \cite{Zhou+:2017} to each 
$\mathbf{h}_i$.  By defining the representation of the whole context
$\mathbf{\overline{h}} = [\vec{\mathbf{h}}_C; \cev{\mathbf{h}}_1]$
(where $C$ is the number of words in the context), the selective
mechanism filters the hidden states $\mathbf{h}_i$ using
$\mathbf{\overline{h}}$ as follows:
\begin{equation*}
    \mathbf{h}'_i = \mathbf{h}_i \odot \sigma(\mathbf{W}_a \mathbf{h}_i 
+ \mathbf{U}_a \mathbf{\overline{h}} + \mathbf{b}_a)
\end{equation*}
where $\odot$ denotes element-wise product. Hereinafter $\mathbf{W}$, 
$\mathbf{U}$ and $\mathbf{b}$ are used to refer to model parameters. The 
intuition behind this procedure is to selectively filter less useful 
elements from the context words.

In the decoder, we embed words $x_t$ in the current line using the 
encoder-shared embedding matrix ($\mathbf{W}_{wrd}$) to produce 
$\mathbf{w}_t$.  In addition to the word embeddings, we also embed the 
characters of a word using embedding matrix $\mathbf{W}_{chr}$ to 
produce $\mathbf{c}_{t,i}$, and feed them to a bidirectional 
(character-level) LSTM:
\begin{align}
\begin{split}
    \vec{\mathbf{u}}_{t,i} &= \text{LSTM}_{f}(\mathbf{c}_{t,i}, 
\vec{\mathbf{u}}_{t,i-1}) \\
    \cev{\mathbf{u}}_{t,i} &= \text{LSTM}_{b}(\mathbf{c}_{t,i}, 
\cev{\mathbf{u}}_{t,i+1})
\end{split}
\label{eqn:char-lstm}
\end{align}
We represent the character encoding of a word by concatenating the last 
forward and first backward hidden states $\mathbf{\overline{u}}_t = 
[\vec{\mathbf{u}}_{t,L}; \cev{\mathbf{u}}_{t,1}]$, where $L$ is the 
length of the word. We incorporate character encodings because they 
provide orthographic information, improve representations of unknown 
words, and are shared with the pentameter model 
(\secref{pm}).\footnote{We initially shared the character encodings with 
the rhyme model as well, but found sub-par performance for the rhyme 
model.  This is perhaps unsurprising, as rhyme and stress are 
qualitatively very different aspects of forms.} The rationale for 
sharing the parameters is that we see word stress and language model 
information as complementary.

Given the word embedding $\mathbf{w}_t$ and character encoding 
$\mathbf{\overline{u}}_t$, we concatenate them together and feed them to 
a unidirectional (word-level) LSTM to produce the decoding states:
\begin{equation}
\mathbf{s}_t = \text{LSTM}([\mathbf{w}_t; \mathbf{\overline{u}}_t], 
\mathbf{s}_{t-1})
\label{eqn:char-word-lm}
\end{equation}

We attend $\mathbf{s}_t$ to encoder hidden states $\mathbf{h}'_i$ and 
compute the weighted sum of $\mathbf{h}'_i$ as follows:
\begin{align*}
    e^t_i &= \mathbf{v}_b^\intercal \tanh(\mathbf{W}_b \mathbf{h}'_i + 
\mathbf{U}_b \mathbf{s}_t + \mathbf{b}_{b}) \\
    \mathbf{a}^t &= \text{softmax}(\mathbf{e}^t) \\
    \mathbf{h}^{*}_t &= \sum_i a^t_i \mathbf{h}'_i
\end{align*}

To combine $\mathbf{s}_t$ and $\mathbf{h}^{*}_t$, we use a gating unit 
similar to a GRU \citep{Cho+:2014,Chung+:2014}: $\mathbf{s}'_t = 
\text{GRU}(\mathbf{s}_t, \mathbf{h}^{*}_t)$.
We then feed $\mathbf{s}'_t$ to a linear layer with softmax activation
to produce the vocabulary distribution (i.e.\
$\text{softmax}(\mathbf{W}_{out} \mathbf{s}'_t + \mathbf{b}_{out})$, and
optimise the model with standard categorical cross-entropy loss. We use
dropout as regularisation \cite{Srivastava+:2014}, and apply it to the
encoder/decoder LSTM outputs and word embedding lookup. The same
regularisation method is used for the pentameter and rhyme models.

As our sonnet data is relatively small for training a neural language 
model (367K words; see \tabref{dataset}), we pre-train word embeddings 
and reduce parameters further by introducing weight-sharing between 
output matrix $\mathbf{W}_{out}$ and embedding matrix $\mathbf{W}_{wrd}$ 
via a projection matrix $\mathbf{W}_{prj}$ 
\cite{Inan+:2016,Paulus+:2017,Press+:2017}:
\begin{equation*}
    \mathbf{W}_{out} = \tanh (\mathbf{W}_{wrd} \mathbf{W}_{prj})
\end{equation*}

\subsection{Pentameter Model}
\label{sec:pm}

This component is designed to capture the alternating iambic stress
pattern.  Given a sonnet line, the pentameter model learns to attend to
the appropriate characters to predict the 10 binary stress symbols
sequentially.\footnote{That is, given the input line \ex{Shall I compare
    thee to a summer's day?} the model is required to output $S^-$ $S^+$
  $S^-$ $S^+$ $S^-$ $S^+$ $S^-$ $S^+$ $S^-$ $S^+$, based on the syllable
  boundaries from \secref{data}.} As punctuation is not pronounced, we
preprocess each sonnet line to remove all punctuation, leaving only
spaces and letters.  Like the language model, the pentameter model is
fashioned as an encoder--decoder network.

In the encoder, we embed the characters using the shared embedding 
matrix $\mathbf{W}_{chr}$ and feed them to the shared bidirectional 
character-level LSTM (\eqnref{char-lstm}) to produce the character 
encodings for the sentence: $\mathbf{u}_j = [\vec{\mathbf{u}}_j; 
\cev{\mathbf{u}}_j]$.  

In the decoder, it attends to the characters to predict the stresses 
sequentially with an LSTM:
\begin{equation*}
    \mathbf{g}_t = \text{LSTM} (\mathbf{u}^*_{t-1}, \mathbf{g}_{t-1})
\end{equation*}
where $\mathbf{u}^*_{t-1}$ is the weighted sum of character encodings 
from the previous time step, produced by an attention network which we 
describe next,\footnote{Initial input ($\mathbf{u}^*_0$) and state 
($\mathbf{g}_0$) is a trainable vector and zero vector respectively.} 
and $\mathbf{g}_t$ is fed to a linear layer with softmax activation to 
compute the stress distribution.

The attention network is designed to focus on stress-producing 
characters, whose positions are monotonically increasing (as stress is 
predicted sequentially).  We first compute $\mu_t$, the mean position of 
focus:
\begin{align*}
    \mu'_t &= \sigma(\mathbf{v}_c^{\intercal}\tanh(\mathbf{W}_c 
\mathbf{g}_t + \mathbf{U}_c \mu_{t-1} + \mathbf{b}_c)) \\
    \mu_t &= M \times \text{min} (\mu'_t+\mu_{t-1}, 1.0)
\end{align*}
where $M$ is the number of characters in the sonnet line.  Given 
$\mu_t$, we can compute the (unnormalised) probability for each 
character position:
\begin{equation*}
p^t_j = \text{exp}\left(\frac{-(j-\mu_t)^{2}}{2T^{2}}\right)
\end{equation*}
where standard deviation $T$ is a hyper-parameter. We incorporate this 
position information  when computing $\mathbf{u}^*_t$:\footnote{Spaces 
are masked out, so they always yield zero attention weights.}
\begin{align*}
    \mathbf{u}'_j &= p^t_j \mathbf{u}_j \\
    d^t_j &= \mathbf{v}^{\intercal}_d \tanh(\mathbf{W}_d \mathbf{u}'_j + 
\mathbf{U}_d \mathbf{g}_t + \mathbf{b}_d)\\
    \mathbf{f}^t &= \text{softmax} (\mathbf{d}^t + \log \mathbf{p}^t) \\
    \mathbf{u}^*_t &= \sum_j b^t_j \mathbf{u}_j
\end{align*}

Intuitively, the attention network incorporates the position information 
at two points, when computing: (1)  $d^t_j$ by weighting the character 
encodings; and (2)  $\mathbf{f}^t$ by adding the position log 
probabilities.  This may appear excessive, but preliminary experiments 
found that this formulation produces the best performance.

In a typical encoder--decoder model, the attended encoder vector 
$\mathbf{u}^*_t$ would be combined with the decoder state $\mathbf{g}_t$ 
to compute the output probability distribution. Doing so, however, would 
result in a zero-loss model as it will quickly learn that it can simply 
ignore $\mathbf{u}^*_t$ to predict the alternating stresses based on 
$\mathbf{g}_t$. For this reason we use only $\mathbf{u}^*_t$ to compute 
the stress probability:
\begin{equation*}
   P(S^-)  = \sigma ( \mathbf{W}_e \mathbf{u}^*_t + b_e )
\end{equation*}
which gives the loss $ \mathcal{L}_{ent} = \sum_t - \log P(S^\star_t)$ 
for the whole sequence, where $S^\star_t$ is the target stress at time 
step $t$.

We find the decoder still has the tendency to attend to the same 
characters, despite the incorporation of position information.  To 
regularise the model further, we introduce two loss penalties: repeat 
and coverage loss.

The repeat loss penalises the model when it attends to previously 
attended characters \cite{See+:2017}, and is computed as follows:
\begin{equation*}
    \mathcal{L}_{rep} = \sum_t \sum_j \min(f^t_j, \sum^{t-1}_{t=1} 
f^t_j)
\end{equation*}

By keeping a sum of attention weights over all previous time steps, we 
penalise the model when it focuses on characters that have non-zero 
history weights.

The repeat loss discourages the model from focussing on the same 
characters, but does not assure that the appropriate characters receive 
attention.  Observing that stresses are aligned with the vowels of a 
syllable, we therefore penalise the model when vowels are ignored:
\begin{equation*}
    \mathcal{L}_{cov} = \sum_{j \in V} \text{ReLU}( C - \sum^{10}_{t=1} 
f^t_j )
\end{equation*}
where $V$ is a set of positions containing vowel characters, and $C$ is 
a hyper-parameter that defines the minimum attention threshold that 
avoids penalty.

To summarise, the pentameter model is optimised with the following loss:
\begin{equation}
    \mathcal{L}_{pm} = \mathcal{L}_{ent} + \alpha \mathcal{L}_{rep} + 
\beta \mathcal{L}_{cov}
\label{eqn:pm-loss}
\end{equation}
where $\alpha$ and $\beta$ are hyper-parameters for weighting the 
additional loss terms.

\subsection{Rhyme Model}
\label{sec:rm}

Two reasons motivate us to learn rhyme in an unsupervised manner: (1) we 
intend to extend the current model to poetry in other languages (which 
may not have pronunciation dictionaries); and (2) the language in our 
\sonnet data is not Modern English, and so contemporary dictionaries may 
not
accurately reflect the rhyme of the data.

Exploiting the fact that rhyme exists in a quatrain, we feed 
sentence-ending word pairs of a quatrain as input to the rhyme model and 
train it to learn how to separate rhyming word pairs from non-rhyming 
ones.  Note that the model does not assume any particular rhyming scheme 
--- it works as long as quatrains have rhyme.

A training example consists of a number of word pairs, generated by 
pairing one target word with 3 other reference words in the quatrain, 
i.e.\ \{$(x_t, x_r), (x_t, x_{r+1}), (x_t, x_{r+2})$\}, where $x_t$ is 
the target word and $x_{r+i}$ are the reference words.\footnote{E.g.\
for the quatrain in \figref{sonnet18}, a training example is 
\{(\ex{day}, \ex{temperate}), (\ex{day}, \ex{may}), (\ex{day}, 
\ex{date})\}.} We assume that in these  
3 pairs there should be one rhyming and 2 non-rhyming pairs. From 
  preliminary experiments we found that we can improve the model by 
introducing additional non-rhyming or negative reference words. Negative 
reference words are sampled uniform randomly from the vocabulary, and 
the number of additional negative words is a hyper-parameter.  

For each word $x$ in the word pairs we embed the characters using the 
shared embedding matrix $\mathbf{W}_{chr}$ and feed them to an LSTM to 
produce the character states ${\mathbf{u}}_{j}$.\footnote{The character 
embeddings are the only shared parameters in this model.}  Unlike the 
language and pentameter models, we use a unidirectional forward LSTM here 
(as rhyme is largely determined by the final characters), and the LSTM 
parameters are not shared. We represent the encoding of the whole word 
by taking the last state $ {\overline{\mathbf{u}}} = {\mathbf{u}}_{L}$, 
where $L$ is the character length of the word.

Given the character encodings, we use a margin-based loss to optimise 
the model:
\begin{align*}
    Q &= \{\text{cos}(\overline{\mathbf{u}}_t, \overline{\mathbf{u}}_r),
    \text{cos}(\overline{\mathbf{u}}_t, \overline{\mathbf{u}}_{r+1}), 
    ...  \} \\
    \mathcal{L}_{rm} &= \max(0, \delta - \text{top}(Q, 1) + 
    \text{top}(Q, 
    2))
\end{align*}
where $\text{top}(Q, k)$ returns the $k$-th largest element in $Q$, and 
$\delta$ is a margin hyper-parameter.

Intuitively, the model is trained to learn a sufficient margin (defined 
by $\delta$) that separates the best pair with \textit{all others}, with 
the second-best being used to quantify \textit{all others}.  This is the 
justification used in the multi-class SVM literature for a similar 
objective \cite{Wang+:2014}.

With this network we can estimate whether two words rhyme by computing 
the cosine similarity score during generation, and resample words as 
necessary to enforce rhyme.

\subsection{Generation Procedure}

We focus on quatrain generation in this work, and so the aim is to
generate 4 lines of poetry.  During generation we feed the hidden state
from the previous time step to the language model's decoder to compute
the vocabulary distribution for the current time step. Words are sampled
using a temperature between 0.6 and 0.8, and they are resampled if the
following set of words is generated: (1) UNK token; (2) non-stopwords
that were generated before;\footnote{We use the NLTK stopword list
  \cite{Bird+:2009}.} (3) any generated words with a frequency
$\geqslant$ 2; (4) the preceding 3 words; and (5) a number of symbols
including parentheses, single and double quotes.\footnote{We add 
  these constraints to prevent the model from being too
  repetitive, in generating the same words.} The first sonnet line is
generated without using any preceding context.

We next describe how to incorporate the pentameter model for generation.  
Given a sonnet line, the pentameter model computes a loss 
$\mathcal{L}_{pm}$ (\eqnref{pm-loss}) that indicates how well the line 
conforms to the iambic pentameter. We first generate 10 candidate lines 
(all initialised with the same hidden state), and then sample one line 
from the candidate lines based on the pentameter loss values 
($\mathcal{L}_{pm}$).  We convert the losses into probabilities by
taking the softmax, and a sentence is sampled with temperature $=0.1$.

To enforce rhyme, we randomly select one of the rhyming schemes 
(\textit{AABB}, \textit{ABAB} or \textit{ABBA}) and resample 
sentence-ending words as necessary.  Given a pair of words, the rhyme 
model produces a cosine similarity score that estimates how well the two 
words rhyme. We resample the second word of a rhyming pair (e.g.\ when 
generating the second \textit{A} in \textit{AABB}) until it produces a 
cosine similarity $\geqslant$ 
0.9.  We also resample the second word of a non-rhyming pair (e.g.  when 
  generating the first \textit{B} in \textit{AABB}) by requiring a 
cosine similarity $\leqslant$ 
0.7.\footnote{Maximum number of resampling steps is capped at 1000.  If 
the threshold is exceeded the model is reset to generate from scratch 
again.}

When generating in the forward direction we can never be sure that any
particular word is the last word of a line, which creates a problem for
resampling to produce good rhymes. This problem is resolved in our model
by reversing the direction of the language model, i.e.\ generating the
last word of each line first. We apply this inversion trick at the word
level (character order of a word is not modified) and only to the
language model; the pentameter model receives the original word order as
input.

\section{Experiments}


We assess our sonnet model in two ways: (1) component evaluation of the
language, pentameter and rhyme models; and (2) poetry generation
evaluation, by crowd workers and an English literature expert. A sample 
of machine-generated sonnets are included in the supplementary material.


We tune the hyper-parameters of the model over the development data 
(optimal configuration in the supplementary material).
Word embeddings are initialised
with pre-trained skip-gram embeddings
\cite{Mikolov+:2013b,Mikolov+:2013c} on the \background dataset, and are
updated during training. For optimisers, we use Adagrad
\cite{Duchi+:2011} for the language model, and Adam \cite{Kingma+:2014}
for the pentameter and rhyme models. We truncate backpropagation through
time after 2 sonnet lines, and train using 30 epochs, resetting the
network weights to the weights from the previous epoch whenever
development loss worsens.


\subsection{Component Evaluation}
\label{sec:component-eval}

\begin{table}[t]
\small
\begin{center}
\begin{adjustbox}{max width=\linewidth}
\begin{tabular}{cccc}
\toprule
\textbf{Model} & \textbf{Ppl} & \textbf{Stress Acc} & \textbf{Rhyme F1} 
\\
\midrule
\lm & 90.13 & -- & -- \\
\lma & 84.23 & -- & -- \\
\lmb & 80.41 & -- & -- \\
\lmbconv & 83.68 & -- & -- \\
\lmbpmrm & 80.22 & 0.74 & 0.91 \\
\hdashline
\stressb & -- & 0.80 & -- \\
\rhymeb & -- & -- & 0.74 \\
\rhymeem & -- & -- & 0.71 \\
\bottomrule
\end{tabular}
\end{adjustbox}
\end{center}
\caption{Component evaluation for the language model (``Ppl'' = perplexity), pentameter model (``Stress Acc''), and rhyme model (``Rhyme 
F1'').  Each number is an average across 10 runs.}
\label{tab:component-eval}
\end{table}

 \begin{figure*}[t]
         \centering
         \begin{subfigure}[b]{0.47\textwidth}
             \includegraphics[width=\textwidth]{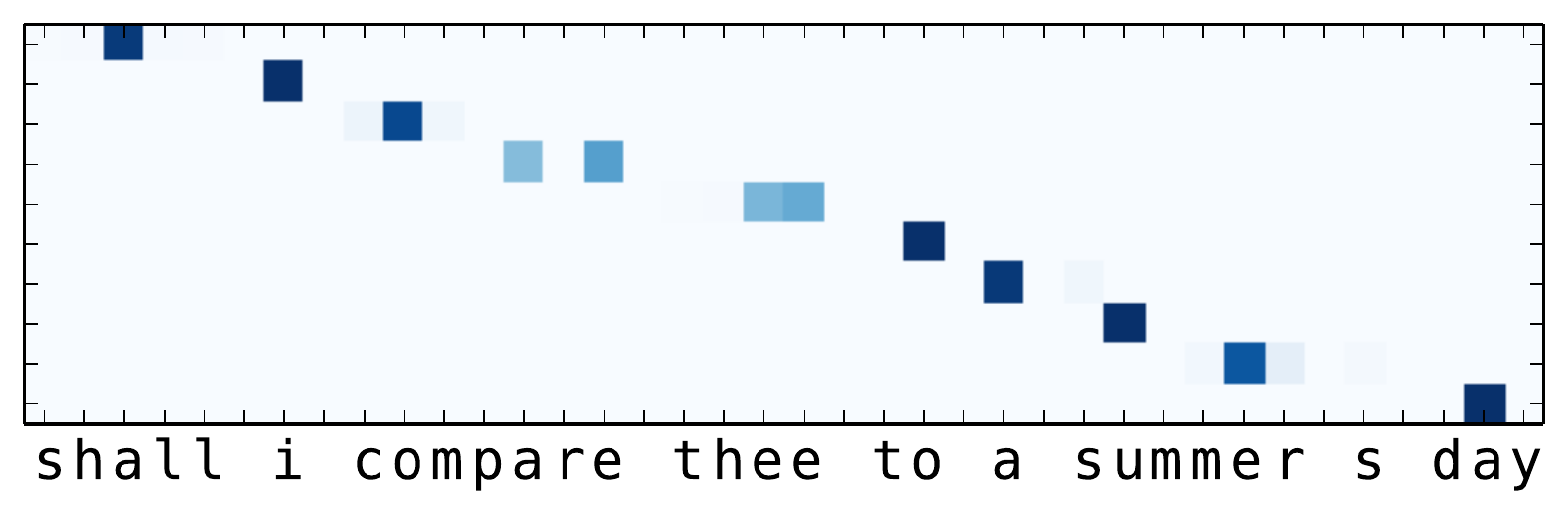}
         \end{subfigure}
         ~
         \begin{subfigure}[b]{0.47\textwidth}
             \includegraphics[width=\textwidth]{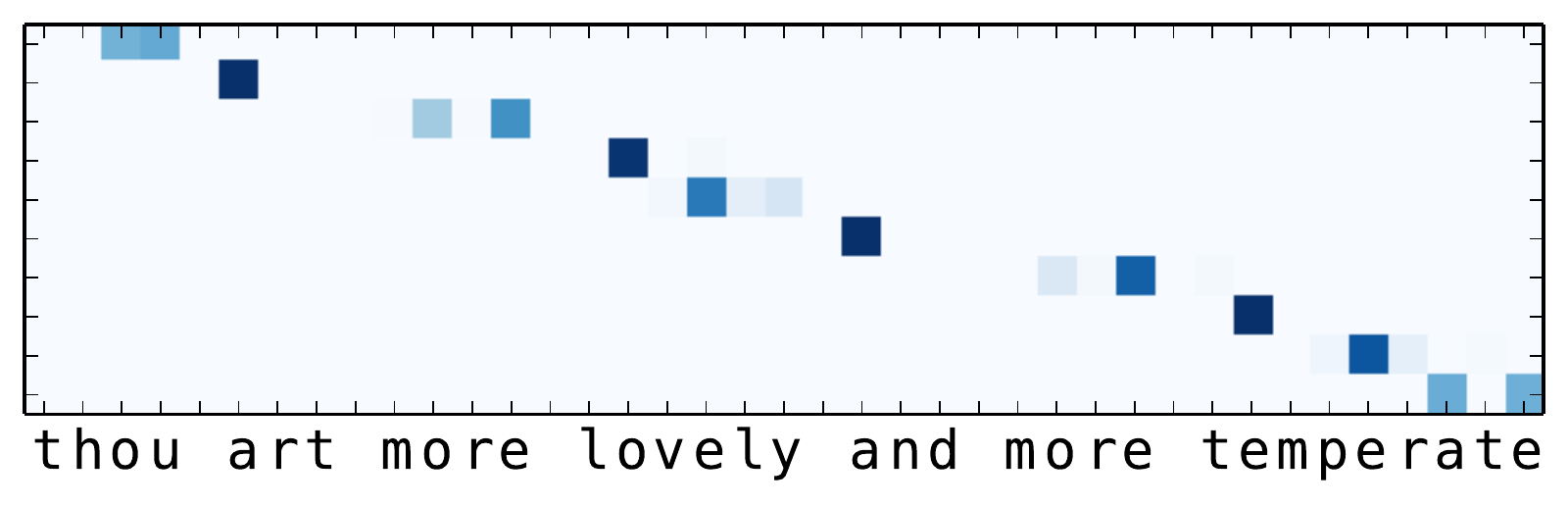}
         \end{subfigure}
         ~
         \begin{subfigure}[b]{0.47\textwidth}
             \includegraphics[width=\textwidth]{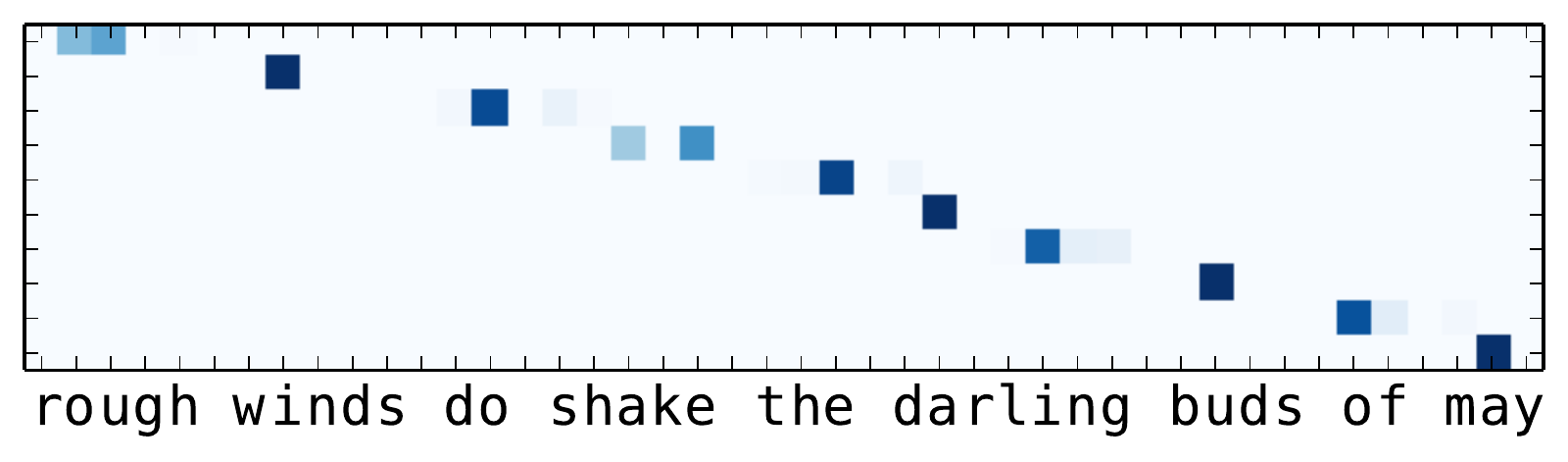}
         \end{subfigure}
         ~
         \begin{subfigure}[b]{0.47\textwidth}
             \includegraphics[width=\textwidth]{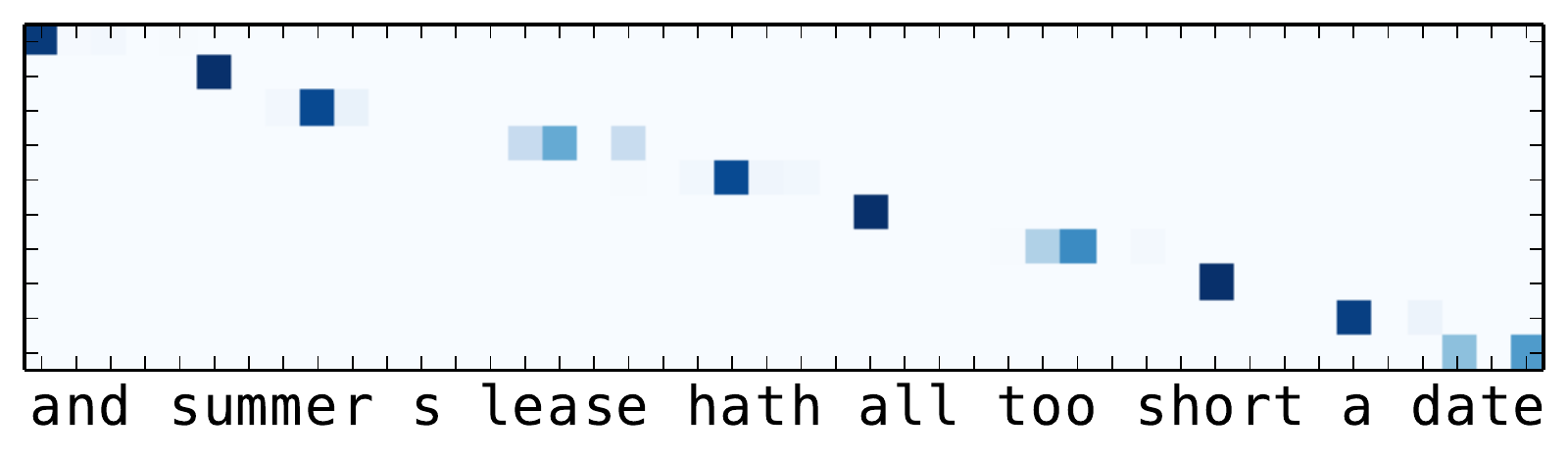}
         \end{subfigure}
         \caption{Character attention weights for the first quatrain of
           Shakespeare's \textit{Sonnet 18}.}
 \label{fig:attention}
 \end{figure*}

\subsubsection{Language Model}

We use standard perplexity for evaluating the language model.  In terms 
of model variants, we have:\footnote{All models use the same 
(applicable) hyper-parameter configurations.}
\begin{itemize}[nosep,leftmargin=1em,labelwidth=*,align=left]
\item \textbf{\lm:} Vanilla LSTM language model;
\item \textbf{\lma:} LSTM language model that incorporates character 
encodings (\eqnref{char-word-lm});
\item \textbf{\lmb:} LSTM language model that incorporates both
character encodings and preceding context;
\item \textbf{\lmbconv:} Similar to \lmb, but preceding context is 
encoded using convolutional networks, inspired by the poetry model of 
\newcite{Zhang+:2014};\footnote{In \newcite{Zhang+:2014}, the authors 
use a series of convolutional networks with a width of 2 words to 
convert 5/7 poetry lines into a fixed size vector; here we use a 
standard convolutional network with max-pooling operation 
\cite{Kim:2014} to process the context.}
\item \textbf{\lmbpmrm:} the full model, with joint training of the language, 
pentameter and rhyme models.
\end{itemize}

Perplexity on the test partition is detailed in 
\tabref{component-eval}. Encouragingly, we see 
that the incorporation of character encodings and preceding context
improves performance substantially, reducing perplexity by almost 10 
points from \lm to \lmb. The inferior performance of \lmbconv compared 
to \lmb demonstrates that our approach of processing context with 
recurrent networks with selective encoding is more effective than 
convolutional networks.  The full model \lmbpmrm, which learns stress 
and rhyme patterns simultaneously, also appears to improve the language 
model slightly.

\subsubsection{Pentameter Model}

To assess the pentameter model, we use the attention weights to predict
stress patterns for words in the test data, and compare them against
stress patterns in the CMU pronunciation
dictionary.\footnote{\url{http://www.speech.cs.cmu.edu/cgi-bin/cmudict}.
  Note that the dictionary provides 3 levels of stresses: 0, 1 and 2; we
  collapse 1 and 2 to $S^+$.} Words that have no coverage or have
non-alternating patterns given by the dictionary are discarded. We use
accuracy as the metric, and a predicted stress pattern is judged to be
correct if it matches any of the dictionary stress patterns.

To extract a stress pattern for a word from the model, we iterate 
through the pentameter (10 time steps), and append the appropriate 
stress (e.g.\ 1st time step $= S^-$) to the word if any of its 
characters receives an attention $\geqslant$ 0.20. 

For the baseline (\stressb) we use the pre-trained weighted finite state
transducer (WFST) provided by
\newcite{Hopkins+:2017}.\footnote{\url{https://github.com/JackHopkins/ACLPoetry}}
The WFST maps a sequence word to a sequence of stresses by assuming each
word has 1--5 stresses and the full word sequence produces iambic
pentameter.  It is trained using the EM algorithm on a sonnet corpus
developed by the authors.

We present stress accuracy in \tabref{component-eval}. \lmbpmrm performs
competitively, and informal inspection reveals that a number of mistakes
are due to dictionary errors.  To understand the predicted stresses  
qualitatively, we display attention heatmaps for the the first quatrain 
of Shakespeare's \textit{Sonnet 18} in \figref{attention}.  The $y$-axis 
represents the ten stresses of the iambic pentameter, and $x$-axis the 
characters of the sonnet line (punctuation removed). The attention 
network appears to perform very well, without any noticeable errors. The 
only minor exception is \ex{lovely} in the second line, where it 
predicts 2 stresses but the second stress focuses incorrectly on the 
character \ex{e} rather than \ex{y}. Additional heatmaps for the full 
sonnet are provided in the supplementary material.

\begin{table}[t]
\begin{center}
\begin{adjustbox}{max width=\linewidth}
\begin{tabular}{r@{\;\;}lr@{\;\;}l}
\toprule
\multicolumn{2}{c}{\textbf{CMU Rhyming Pairs}} & 
\multicolumn{2}{c}{\textbf{CMU Non-Rhyming Pairs}} \\
\cmidrule(lr){1-2}
\cmidrule(lr){3-4}
Word Pair & Cos & Word Pair & Cos \\
\midrule
\pair{endeavour}{never} & 0.028 & \pair{blood}{stood} & 1.000 \\
\pair{nowhere}{compare} & 0.098 & \pair{mood}{stood} & 1.000 \\
\pair{supply}{sigh} & 0.164 & \pair{overgrown}{frown} & 1.000 \\
\pair{sky}{high} & 0.164 & \pair{understood}{food} & 1.000 \\
\pair{me}{maybe} & 0.165 & \pair{brood}{wood} & 1.000 \\
\pair{cursed}{burst} & 0.172 & \pair{rove}{love} & 0.999 \\
\pair{weigh}{way} & 0.200 & \pair{sire}{ire} & 0.999 \\
\pair{royally}{we} & 0.217 & \pair{moves}{shoves} & 0.998 \\
\pair{use}{juice} & 0.402 & \pair{afraid}{said} & 0.998 \\
\pair{dim}{limb} & 0.497 & \pair{queen}{been} & 0.996 \\

\bottomrule

\end{tabular}
\end{adjustbox}
\end{center}
\caption{Rhyming errors produced by the model. Examples on the left 
(right) side are rhyming (non-rhyming) word pairs --- determined using 
the CMU dictionary --- that have low (high) cosine similarity.  ``Cos'' 
denote the system predicted cosine similarity for the word pair.  }
\label{tab:rhyme}
\end{table}

\subsubsection{Rhyme Model}

We follow a similar approach to evaluate the rhyme model against the
CMU dictionary, but score based on F1 score.  Word pairs that are not
included in the dictionary are discarded. Rhyme is determined by
extracting the final stressed phoneme for
the paired words, and testing if their phoneme patterns match.

We predict rhyme for a word pair by feeding them to the rhyme model and
computing cosine similarity; if a word pair is assigned a score 
$\geqslant$
0.8,\footnote{0.8 is empirically found to be the best threshold based on 
   development data.} it is considered to rhyme.  As a baseline 
(\rhymeb), we first
extract for each word the last vowel and all following consonants, and
predict a word pair as rhyming if their extracted sequences match. The
extracted sequence can be interpreted as a proxy for the last syllable
of a word.

\newcite{Reddy+:2011} propose an unsupervised model for learning rhyme 
schemes in poems via EM. There are two latent variables: $\phi$ 
specifies the distribution of rhyme schemes, and $\theta$ defines the 
pairwise rhyme strength between two words. The model's objective is to 
maximise poem likelihood over all possible rhyme scheme assignments 
under the latent variables $\phi$ and $\theta$. We train this model 
(\rhymeem) on our data\footnote{We use the original authors' 
implementation: \url{https://github.com/jvamvas/rhymediscovery}.} and 
use the learnt $\theta$ to decide whether two words rhyme.\footnote{A 
word pair is judged to rhyme if $\theta_{w_1, w_2} \geqslant 0.02$; the 
threshold (0.02) is selected based on development performance.} 

\tabref{component-eval} details the rhyming results. The rhyme model
performs very strongly at F1 $>$ 0.90, well above both baselines.  
\rhymeem performs poorly because it operates at the word level (i.e.\ it 
ignores character/orthographic information) and hence does not 
generalise well to unseen words and word pairs.\footnote{Word pairs that 
did not co-occur in a poem in the training data have rhyme strength of 
zero.}

To better understand the errors qualitatively, we present a list of word 
pairs with their predicted cosine similarity in \tabref{rhyme}. Examples 
on the left side are rhyming word pairs as determined by the CMU 
dictionary; right are non-rhyming pairs. Looking at the rhyming word 
pairs (left), it appears that these words tend not to share any 
word-ending characters.  For the non-rhyming pairs, we spot several CMU 
errors: 
\pair{sire}{ire} and \pair{queen}{been} clearly rhyme.

\subsection{Generation Evaluation}
\label{sec:generation-eval}


\subsubsection{Crowdworker Evaluation}
\label{sec:crowd-eval}

Following \newcite{Hopkins+:2017}, we present a pair of quatrains (one
machine-generated and one human-written, in random order) to crowd
workers on CrowdFlower, and ask
them to guess which is the human-written poem. Generation quality is
estimated by computing the accuracy of workers at correctly identifying
the human-written poem (with lower values indicate better results for
the model).

We generate 50 quatrains each for \lm, \lmb and \lmbpmrm (150 in total),
and as a control, generate 30 quatrains with \lm trained for one
epoch. An equal number of human-written quatrains was sampled from the
training partition.  A HIT contained 5 pairs of poems (of which one is a
control), and workers were paid \$0.05 for each HIT. Workers who failed
to identify the human-written poem in the control pair reliably (minimum
accuracy $=$ 70\%) were removed by CrowdFlower automatically, and
they were restricted to do a maximum of 3 HITs. To dissuade workers from 
using search engines to identify real poems, we presented the quatrains 
as images.


Accuracy is presented in \tabref{crowd-results}. We see a steady
decrease in accuracy (= improvement in model quality) from \lm to \lmb
to \lmbpmrm, indicating that each model generates quatrains that are
less distinguishable from human-written ones.  Based on the suspicion
that workers were using rhyme to judge the poems, we tested a second
model, \lmbrm, which is the full model without the pentameter
component. We found identical accuracy (0.532), confirming our suspicion
that crowd workers depend on only rhyme in their judgements.  These
observations demonstrate that meter is largely ignored by lay persons in
poetry evaluation.

%
%
%
%

\begin{table}[t]
\small
\begin{center}
\begin{tabular}{cc}
\toprule
\textbf{Model} & \textbf{Accuracy} \\
\midrule
\lm & 0.742 \\
\lmb & 0.672 \\
\lmbpmrm & 0.532 \\
\hdashline
\lmbrm & 0.532 \\
\bottomrule
\end{tabular}
\end{center}
\caption{Crowdworker accuracy performance.}
\label{tab:crowd-results}
\end{table}

\begin{table}[t]
\begin{center}
\begin{adjustbox}{max width=\linewidth}
\begin{tabular}{cc@{}cc@{}cc@{}cc@{}c}
\toprule
\textbf{Model} & \multicolumn{2}{c}{\textbf{Meter}} & 
\multicolumn{2}{c}{\textbf{Rhyme}} & \multicolumn{2}{c}{\textbf{Read.}} 
& \multicolumn{2}{c}{\textbf{Emotion}} \\
\midrule
\lm & 4.00 & \sd{0.73} & 1.57 & \sd{0.67} & 2.77 & \sd{0.67} & 
2.73 & \sd{0.51} \\
\lmb & 4.07 & \sd{1.03} & 1.53 & \sd{0.88} & 3.10 & \sd{1.04} & 2.93 & 
\sd{0.93} \\
\lmbpmrm & 4.10 & \sd{0.91} & 4.43 & \sd{0.56} & 2.70 & \sd{0.69} & 2.90 
& \sd{0.79} \\
\hdashline
\human & 3.87 & \sd{1.12} & 4.10 & \sd{1.35} & 4.80 & \sd{0.48} & 4.37 & 
\sd{0.71} \\
\bottomrule
\end{tabular}
\end{adjustbox}
\end{center}
\caption{Expert mean and standard deviation ratings on several aspects 
of the generated quatrains.}
\label{tab:expert-evaluation}
\end{table}

\subsubsection{Expert Judgement}
\label{sec:expert-eval}

To better understand the qualitative aspects of our generated quatrains,
we asked an English literature expert (a Professor of English literature
at a major English-speaking university; the last author of this paper) to directly rate 4 aspects:
meter, rhyme, readability and emotion (i.e.\ amount of emotion the poem
evokes). All are rated on an ordinal scale between 1 to 5 (1 $=$ worst;
5 $=$ best). In total, 120 quatrains were annotated, 30 each for \lm,
\lmb, \lmbpmrm, and human-written poems (\human). The expert was blind
to the source of each poem. The mean and standard deviation of the
ratings are presented in \tabref{expert-evaluation}.

We found that our full model has the highest ratings for both rhyme and
meter, even higher than human poets. This might seem surprising, but in
fact it is well established that real poets regularly break rules of
form to create other effects \cite{Adams97}. Despite excellent form, the 
output of our model can easily be distinguished from human-written
poetry due to its lower emotional impact and readability. In
particular, there is evidence here that our focus on form actually hurts 
the readability of the resulting poems, relative even to the simpler
language models. Another surprise is how well simple language models do
in terms of their grasp of meter: in this expert evaluation, we see only
marginal benefit as we increase the sophistication of the
model. Taken as a whole, this evaluation suggests that future research
should look beyond forms, towards the substance of good poetry.



%
%
%

\section{Conclusion}

We propose a joint model of language, meter and rhyme that captures
language and form for modelling sonnets. We provide quantitative
analyses for each component, and assess the quality of generated poems
using judgements from crowdworkers and a literature expert.  Our
research reveals that vanilla LSTM language model captures meter
implicitly, and our proposed rhyme model performs exceptionally well.
Machine-generated generated poems, however, still underperform in terms
of readability and emotion.


\bibliography{strings,ref,papers}
\bibliographystyle{acl_natbib}

\end{document}